\documentclass[10pt,twocolumn,letterpaper]{article}

\usepackage{cvpr}
\usepackage{times}
\usepackage{epsfig}
\usepackage{graphicx}
\usepackage{amsmath}
\usepackage{amssymb}
\usepackage{mathtools}
\usepackage{url}
\usepackage{listings}
\usepackage{framed}
\usepackage{multirow}
\usepackage{xcolor}
\usepackage{mdwlist}
\usepackage{caption}
\usepackage{subcaption}

\newcommand{\SotA} {state-of-the-art }
\newcommand{\dropin} {DropIn }
\newcommand{\dropinNS} {DropIn}

\usepackage[pagebackref=true,breaklinks=true,letterpaper=true,colorlinks,bookmarks=false]{hyperref}

\cvprfinalcopy 

\def\cvprPaperID{495} 

\begin{document}

\title{Gradual DropIn of Layers to Train Very Deep Neural Networks}

\author{
Leslie N. Smith\\
Naval Research Laboratory\\
{\tt\small leslie.smith@nrl.navy.mil}\\
\and
Emily M. Hand\\
University of Maryland\\
{\tt\small emhand@cs.umd.edu}
\and
Timothy Doster\\
Naval Research Laboratory\\
{\tt\small timothy.doster@nrl.navy.mil}\\
}

\maketitle

\begin{abstract}

We introduce the concept of dynamically growing a neural network during training. In particular, an untrainable deep network starts as a trainable shallow network and newly added layers are slowly, organically added during training, thereby increasing the network's depth. This is accomplished by a new layer, which we call \dropinNS. The \dropin layer starts by passing the output from a previous layer (effectively skipping over the newly added layers), then increasingly including units from the new layers for both feedforward and backpropagation. We show that deep networks, which are untrainable with conventional methods, will converge with \dropin layers interspersed in the architecture. In addition, we demonstrate that \dropin provides regularization during training in an analogous way as dropout. Experiments are described with the MNIST dataset and various expanded LeNet architectures, CIFAR-10 dataset with its architecture expanded from 3 to 11 layers, and on the ImageNet dataset with the AlexNet architecture expanded to 13 layers and the VGG 16-layer architecture.

\end{abstract}

\section{Introduction}

\label{sec:intro}

Over the past few years, \SotA results for image recognition \cite{Alexnet12,  simonyan2014very,szegedy2014going}, object detection \cite{girshick2014rich}, face recognition \cite{taigman2014deepface}, speech recognition \cite{graves2014towards}, machine translation \cite{sutskever2014sequence}, image caption generation \cite{vinyals2014show}, driverless car technology \cite{huval2015empirical}, and other applications \cite{lecun2015deep} have required increasingly deeper neural networks. 

Network depth refers to the number of layers in the architecture. 
It is well known that adding layers to neural networks makes them more expressive \cite{montufar2014number}. 
Each year, the Imagenet Challenge \cite{Imagenet15} is held in which teams are expected, given an image, to detect, localize, or recognize an object in the image. 
Deep convolutional neural networks (CNNs) have dominated the competition since Krizhevsky \etal won in 2012 \cite{Alexnet12}, and each year since, the winner of the competition used a deeper network than the previous year's winner \cite{Imagenet15, simonyan2014very, szegedy2014going}.

However, training a very deep network is a difficult and open research problem \cite{erhan2009difficulty, glorot2010understanding, srivastava2015training}. 
It is difficult to train very deep networks because the error norm during backpropagation can grow or vanish exponentially.
In addition, very large training datasets are necessary when the network has millions of weights.

Here we suggest a dynamic architecture that grows during the training process and allows for the training of very deep networks.
We illustrate this with our \dropin layer, where new layers are skipped at the start of the training, as though they were not present.
This allows the weights of the included layers to start converging.
Over a number of iterations the \dropin layer increasingly includes activations from the inserted layers, which gradually trains the weights in theses added layers. 

\dropin follows the philosophy embedded within curriculum learning \cite{bengio2009curriculum}.
With curriculum learning one starts with an easier problem and incrementally increases the difficulty.
Here too, one starts training a shallow architecture and after convergence begins, \dropin incrementally modifies the architecture to slowly include units from the new layers.

In addition, \dropin can be used in a mode analogous to dropout \cite{srivastava2014dropout} for the regularization of a deep neural network during training.  
Instead of setting random activations to zero, as is done in dropout, \dropin sets these activations to the activations from a previous layer.
We demonstrate that the ``noise'' from mixing the activations from previous layers provides regularization during training. 
In addition, both \dropin and dropout can be viewed as training a large collection of networks with varied architectures and extensive weight sharing.

The contributions of this paper are:
\begin{enumerate*}
  \vspace{-10pt}
\item A dynamic architecture that can grow during training.  
\item The details of a \dropin layer for enabling the training of very deep networks and for regularization during training.  
\item Examples of successfully training deep architectures that cannot be trained with conventional methods on MNIST, CIFAR-10, and ImageNet.
\end{enumerate*}


\section{Related work}
Methods for training very deep networks  have centered on initialization of the network weights or developing new architectures and \dropin is in the latter category.  

\subsection{Initialization of network weights}
Sutskever \etal \cite{sutskever2013importance} investigate the difficulty in training deep networks and conclude that both proper initialization and momentum are necessary.  Glorot and Bengio \cite{glorot2010understanding} recommend an initialization method called \textit{normalized initialization} to allow the training of deep networks.  
He \etal \cite{he2015delving} recently improved  upon the ``normalized initialization'' method by changing the distribution to take into account ReLU layers.  

Hinton \etal \cite{hinton2006fast} proposed first training layer by layer in an unsupervised fashion so that a transformed version of the input could be realized.  
Erhan  \cite{erhan2009difficulty} later characterized the mathematics of the unsupervised pre-training and offered an explanation for its success.   

Sussillo and Abbott \cite{sussillo2015random} suggest an initialization scheme called \textit{Random Walk Initialization} based on scaling the initial random matrices correctly.  
By multiplying the error gradient by a correctly scaled random matrix at each layer, an unbiased  random walk is formed.  
This is one of only a few papers that show the results of experiments with networks consisting of hundreds of layers.  

\subsection{Developing new architecture}

Raiko, \etal \cite{raiko2012deep} introduce the concept of skip connections by adding a linear transformation to the usual non-linear transformation of the input to a unit.  
Skip connections separate the linear and non-linear portions of the activations and allow the linear part to ``skip'' to higher layers.
This is similar to \dropin in some ways, but the purpose of \dropin differs from that of skip connections, and \dropin does not need to learn any parameters.

Romero \etal \cite{romero2014fitnets} suggest training a thin, deep student network (called a \textit{fitnet}) from a larger but shallower teacher network.  
The authors accomplish this by utilizing the output of the teacher's hidden layers as a hint for the student's hidden layers.

Srivastava \etal \cite{srivastava2015highway, srivastava2015training} propose a new architecture, which they named \textit{Highway Networks}, where the output of a layer's neuron contains a combination of the input and the output. 
Highway networks use carry gates inspired by long short-term memory (LSTM) recurrent neural networks (RNNs) to regulate how much of the input is carried to the next layer.  
The authors 
demonstrate that their structure permits training networks of hundreds of layers (up to 900 layers) \cite{srivastava2015highway, srivastava2015training}. 
These new parameters are learned along with the other parameters of the network.   
Zhang \etal \cite{Zhang2015highway} applied highway networks to LSTM recurrent neural networks.
\dropin is a simpler approach than highway networks as it does not contain gate parameters that need to be learned.

Breuel \cite{breuel2015possible} discusses a dynamic network that he describes as a biologically plausible ``reconfigurable'' network.
In this network different units are weighted dynamically to produce different configurations.  This allows a single network to perform multiple tasks.
\dropin represents a different type of dynamic network that grows during training rather than reconfigures for each task.

\subsection{Regularization during training}
The well-known dropout \cite{hinton2012improving, srivastava2014dropout} method is an effective means to improve the training  of deep neural networks.   
During training dropout randomly zeros a neuron's output activation with a probability $p$, called the \textit{dropout ratio},  so that the network cannot rely on a particular configuration.
This reduces overfitting to the training data and the resulting network is more robust and better generalizes to unseen data.  
While dropout ``samples from an exponential number of different `thinned' networks'' \cite{srivastava2014dropout}, \dropin samples from an exponential number of different thinner and shallower sub-networks.
Like dropout, \dropin randomly changes the configuration so that the network cannot rely on a particular configuration. 

Baldi and Sadowski \cite{baldi2013understanding} provide a theoretical basis for understanding dropout,  demonstrating that dropout regulates the training and prevents overfitting by approximating an average of a  large ensemble of networks.  
A similar theoretical understanding (and benefits) can also apply to \dropinNS.



\begin{figure}[tb]
\begin{center}
   \includegraphics[width=0.9\linewidth]{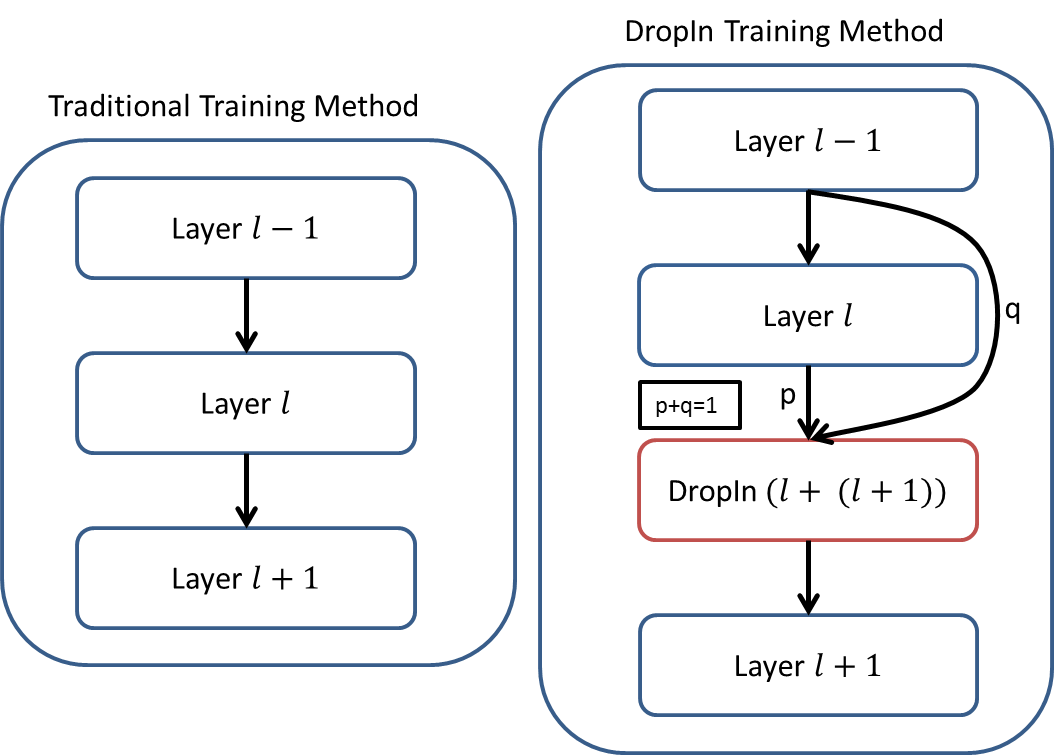}
\end{center}
  \vspace{-10pt}
   \caption{Diagram of traditional vs \dropin training method.  The \dropin method sends activations from Layer $ {\ell} $ to Layer $ {\ell+1} $ with a ratio $p$ and from Layer $ {\ell-1} $ to Layer $ {\ell+1}$ (thus skipping Layer $\ell$) with a ratio $q=1-p$.
}
\label{fig:diagram}
  \vspace{-10pt}
\end{figure}


\section{\dropin method}
In this section we provide a mathematical basis for \dropin as well as some implementation details.

\subsection{Model description}
\label{sec:model}
There are two modes of running \dropinNS: first to gradually include skipped layers, which we refer to as \textit{gradual \dropinNS}, and second as a regularizer, which we named \textit{regularizing \dropinNS}.
Figure \ref{fig:diagram} provides a visual reference as to how the \dropin unit works.

Gradual \dropin initially passes on only the activations from the previous layer, effectively skipping the new layers.  
For each iteration number, $\tau$, the ratio $ p $ is computed as $ p = \tau / d$ for \dropin length $d$, which is the number of iterations over which $q = 1 - p$ reduces from 1 to 0. 
Then the number of activations copied from layer $ {\ell-1} $ drops as  $ q \times n = (1 - p) \times n $, where $n$ is the total number of activations in the layer ${\ell-1}$.  
The remaining activations are accepted from the new layer $ {\ell} $ and backpropagation trains the weights of these newly added units.

For regularizing \dropinNS, the \dropin probability ratio $p$ is set to a static value in $[0,1]$.
In this case, \dropin works analogously with dropout but instead of setting values to zero, they are set to the activations of a previous layer (e.g., $ {\ell-1} $).
The choice of which activations come from which layer is done in an evolving random fashion each iteration.

We follow the notation in the dropout paper \cite{srivastava2014dropout} to show this more formally.
Namely, we start with a neural network composed of some number of layers, $L$, where $ \ell \in [1, 2, \dots, L] $ is the layer index.
Also, $ \mathbf{y}^{(\ell)} $ represents the vector of outputs from layer $ {\ell} $ and is the input to the next layer $ {\ell+1} $.  Let $ \mathbf{x} $ be the data input to the first layer. 
In addition, $ \mathbf{W}^{(\ell)} $ and $ \mathbf{b}^{(\ell)} $ are the weights and biases at layer $ \ell $.  
To allow us to track the evolving nature of the network, we include the training iteration number, $\tau$, and the layer's unit index number, $\lambda^{(\ell)}$.

The first equation for gradual \dropin is a vector of zeros then ones, which is designated as: 
\begin{eqnarray}
\mathbf{r}^{(\ell)}(\tau,\lambda^{(\ell)}) =\begin{cases} 0 &\lambda^{(\ell)}< q \times n \\
1 & \mbox{otherwise}.
\end{cases}\end{eqnarray}
For regularizing \dropinNS, the  equation for $ \mathbf{r}^{(\ell)}(\tau, \lambda^{(\ell)})$ with a probability ratio $ p $ is:
\begin{equation}
\mathbf{r}^{(l)}(\tau,\cdot) \sim Bernoulli(p), 
\end{equation}
i.e.,  a 0-1 vector where each value is distributed as a Bernoulli random variable with probability $p$.  

Once $ \mathbf{r} $  is set, the remaining equations (dropping $\tau$ and $\lambda^{(\ell)}$ for simplicity) are the same for both modes -- namely for layer $ {\ell+1}$:
\begin{equation}
\mathbf{\tilde{y}}^{(\ell)} = \mathbf{r}^{(\ell)} \times \mathbf{y}^{(\ell)}  \\
\end{equation}
\begin{equation}
z^{(\ell+1)}_i  = \mathbf{w}^{(\ell+1)}_i \mathbf{\tilde{y}}^{(\ell)} + b_i^{(\ell+1)} \\
\end{equation}
\begin{equation}
\mathbf{y}^{(\ell+1)} = f ( z_i^{(\ell)} ) + ( 1 - \mathbf{r}^{(\ell)} )  \mathbf{y}^{(\hat{\ell})},
\end{equation}
where $ \hat{\ell} $ is any layer less than layer $ {\ell+1}$.
These equations are similar to those for dropout, except instead of some of the outputs being zero, they are set to the values from a previous layer, $ \mathbf{y}^{(\hat{\ell})} $.

\subsection{Implementation}
\label{sec:code}

We implemented our method in Caffe \cite{Caffe14} by creating a new layer called \dropinNS.
The parameters for the \dropin layer include a $dropin\_ratio$, $ q = 1 - p$ (Figure \ref{fig:diagram}), and a $dropin\_length$, $d$, as described in Section \ref{sec:model}.

\dropin requires that the size of both the new layer and the previous layer be the same.
Hence, we also implemented a Resize layer to allow reshaping a layer's output to a user-specified size.
The Resize layer modifies its input, which is $ \mathbf{y}^{(\hat{\ell})},$ into a user-specified height, width, and number of channels/filters. 
The Resize layer allows \dropin to work with any two layers, even when the sizes of $ \mathbf{y}^{(\ell)}$ and $ \mathbf{y}^{(\hat{\ell})} $ are different.


\section{Experiments}
\label{sec:experiments}

The purpose of this section is to demonstrate the effectiveness of \dropin on several standard datasets but with  architectures that are not trainable with standard methods.
No attempt was made to optimize the architecture or hyper-parameters for higher accuracy because our main objective was to show that a deep architecture that will not converge without \dropinNS, will converge with it.
However, the results in Sections  \ref{sec:alexnet} and \ref{sec:vgg}  also demonstrate an increase in accuracy by using a deeper network for Imagenet.

All of the following experiments were run with Caffe (downloaded August 31, 2015) using CUDA 7.0 and Nvidia's CuDNN.  
These experiments were run on a 64 node cluster with 8 Nvidia Titan Black GPUs, 128 GB memory, and dual Intel Xenon E5-2620 v2 CPUs per node.  

The following subsections depict, in table form, the structure of several networks.  We use the naming convention \{layer type\}\{layer number\}-\{number of outputs\}(filter size).  
For example, conv1\_2-32$(5\times 5)$ represents a convolutional layer numbered 1\_2 with 32 outputs and filters sized $5\times 5$.  
\dropin layers are denoted as dropin $(\ell + (\ell + 1))$, as depicted in Figure \ref{fig:diagram}.

In Section \ref{sec:mnist} we show that very deep networks are trainable when using gradual \dropin on expanded LeNet with MNIST data.  In Section \ref{sec:cifar} we show the effect of \dropin length on training accuracy for expanded CIFAR-10 network and that a small performance gain is possible with added layers.
In Section \ref{sec:alexnet} we show that an expanded AlexNet architecture increases accuracy and is trainable only with gradual \dropin.  
In Section \ref{sec:vgg} the VGG16 network is trained using gradual \dropin without the need to transfer weights from a shallower network.  

\begin{table}[htb]
\begin{center}
  \begin{tabular}{| c | c |}
    \hline
      LeNet & LeNet(2N) {\color{red}+ \dropin} \\ \hline
    
     \multicolumn{2}{|c|}{data $(28\times 28 )$} \\ \hline
     conv1\_1-20$(5\times 5)$ & conv1\_1-20$(5\times 5)$ \\ 
        & conv1\_2-20$(3\times 3)$ \\ 
       & {\color{red}dropin (1\_1 + 1\_2)} \\
        & conv1\_3-20$(3\times 3)$ \\ 
       & {\color{red}dropin (1\_2 + 1\_3)} \\ 
         & $\vdots$ \\ 
        & conv1\_N-20$(3\times 3)$ \\ 
       & {\color{red}dropin (1\_(N-1) + 1\_N)} \\ \hline
	\multicolumn{2}{|c|}{maxpool$(2\times2)$} \\ \hline

     conv2\_1-50$(5\times 5)$ & conv2\_1-50$(5\times 5)$ \\ 
       & conv2\_2-50$(3\times 3)$ \\ 
      & {\color{red}dropin (2\_1 + 2\_2)} \\
      & conv2\_3-50$(3\times 3)$ \\ 
      & {\color{red}dropin (2\_2 + 2\_3)} \\ 
       & $\vdots$ \\ 
      & conv2\_N-50$(3\times 3)$ \\ 
      & {\color{red}dropin (2\_(N-1) + 2\_N)} \\ \hline
\multicolumn{2}{|c|}{maxpool$(2\times2)$} \\ \hline
\multicolumn{2}{|c|}{fc3-500} \\ \hline
\multicolumn{2}{|c|}{fc4-10} \\ \hline
\multicolumn{2}{|c|}{soft-max} \\ \hline
  \end{tabular}
  \caption{Network architecture for LeNet and LeNet(2N)+ \dropinNS.  }
  \label{tab:lenet_N}
\end{center}
  \vspace{-20pt}
\end{table}

\begin{figure}[tbh]
\begin{center}
   \includegraphics[width=1.0\linewidth]{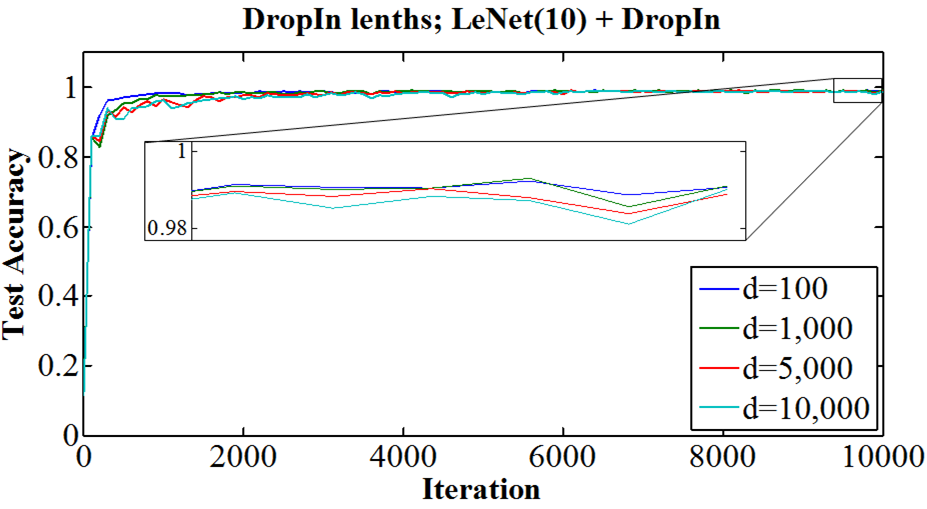}
\end{center}
  \vspace{-10pt}
   \caption{Classification accuracy while training LeNet(10) + \dropin architecture with MNIST data. Curves represent different \dropin lengths, $d$.  (Best viewed in color)  
}
\label{fig:lenet_10}
  \vspace{-10pt}
\end{figure}

\begin{figure}[tbh]
\begin{center}
\begin{subfigure}[b]{0.5\textwidth}
        \includegraphics[width=\textwidth]{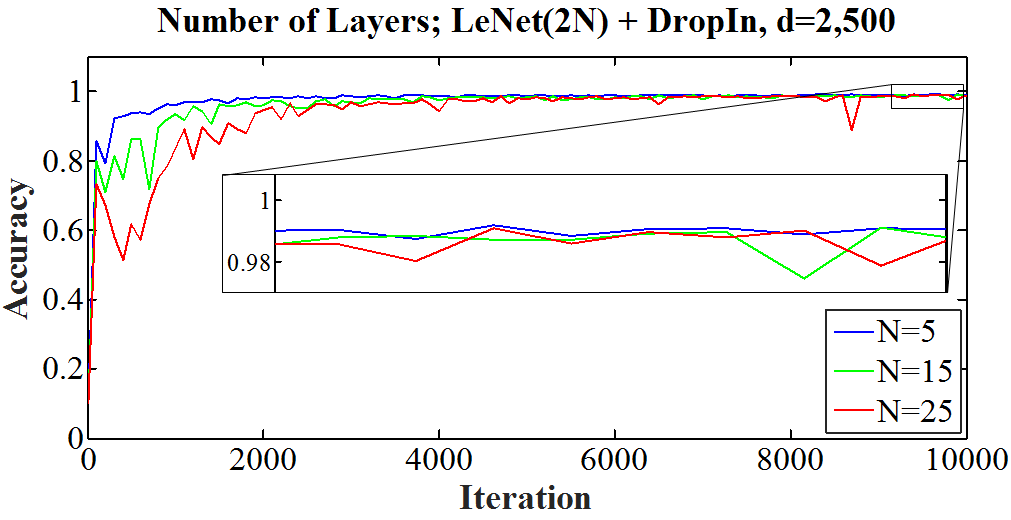}
    \end{subfigure}\\
    \begin{subfigure}[b]{0.5\textwidth}
        \includegraphics[width=\textwidth]{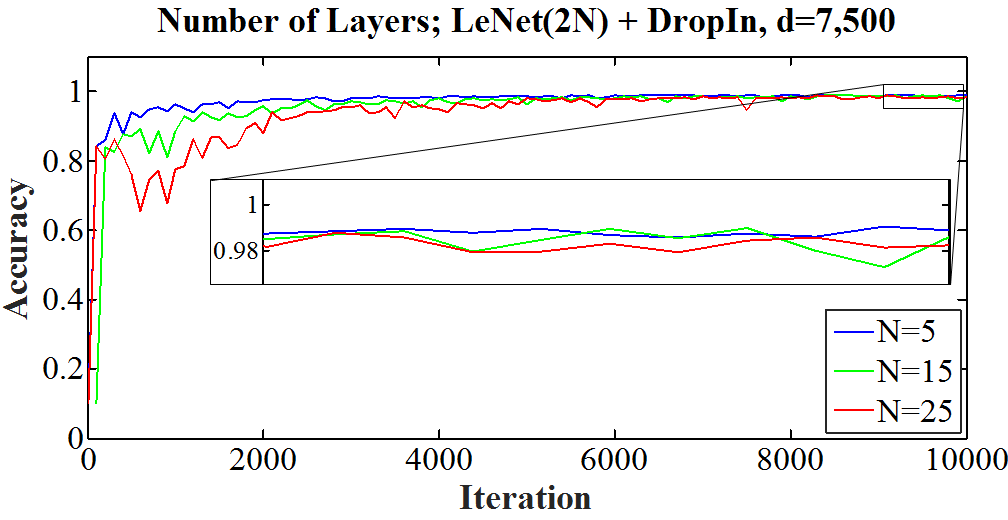}
    \end{subfigure}
\end{center}
 \vspace{-10pt}
   \caption{Classification accuracy while training LeNet(2N) + \dropinNS, for $N=5,15,25$ with MNIST data. Curves represent different network depths. (Best viewed in color)  
}
\label{fig:lenet_droplens}
  \vspace{-10pt}
\end{figure}


\subsection{MNIST} \label{sec:mnist}
This dataset consists of 70,000 grey-scale images with a resolution of 28x28\footnote{\url{http://yann.lecun.com/exdb/mnist/}}.  
Of these, 60,000 are for training and 10,000 are for testing.  
There are ten classes, each a different handwritten digit from zero to nine, with 7,000 images per class.  

The standard network architecture for the classification of MNIST, provided in the Caffe package, is the 4-layer LeNet consisting of 2 convolutional/max-pooling layers followed by 2 fully-connected layers (see the first column of Table \ref{tab:lenet_N} for details).  
Inspired by the work in \cite{srivastava2015training}, we increased the number of convolutional layers from two to 2N, which we denote as LeNet(2N).  
These added layers (as seen in the second column of Table \ref{tab:lenet_N}, minus the \dropin layers shown in red) learned a $3\times 3$ convolution filter but did not change the size of the outputs.  
We then added \dropin layers between each of the convolutional layers (as seen in the second column of Table \ref{tab:lenet_N}) and called this network LeNet(2N) + \dropinNS.

We first looked at $N=5$ and created LeNet(10) and LeNet(10) + \dropin architectures.   LeNet(10) did not converge in the standard training time of 10,000 iterations given multiple realizations of the training process.  
However, utilizing \dropin units we were able to have LeNet(2N) + \dropin  converge 10,000 iterations with the same hyper-parameters.  
In Figure \ref{fig:lenet_10} we show results for several different \dropin lengths for this network.  These different lengths indicate the robustness of the \dropin length for simpler networks and that, in general, shorter \dropin lengths provide marginally better results.
We note for this case that the added layers do not increase the overall accuracy of the network, as the MNIST data is quite simple compared with other classification tasks; the added layers do not provide any extra differentiation power.

We now look at how the number of layers affects the training with \dropinNS.  
In Figure \ref{fig:lenet_droplens} there are two different plots, one with \dropin length of 2,500 iterations and the other with \dropin length of 7,500 iterations.  For each plot we present three different networks with 10, 30, and 50, convolutional layers (equating to N=5, 15, 25).  
For both \dropin lengths and all three network depths, the gradual \dropin method allowed the networks to converge.  
The deeper networks require a greater number of iterations to reach the same level of accuracy as the shallower networks, which is to be expected as they have a greater number of weights to train.  We also see that networks converge  more quickly with the shorter \dropin length,  indicating that shorter \dropin lengths are desirable.


\begin{table}[htb]
\begin{center}
  \begin{tabular}{| c | c |}
    \hline
     CIFAR-10 &  CIFAR-10(11 layers) {\color{red}+ \dropin}   \\ \hline
    
     \multicolumn{2}{|c|}{data $(32\times32\times3)$} \\ \hline
     conv1-32$(5\times 5)$ & conv1\_1-32$(5\times 5)$ + LRN \\ 
     maxpool$(2\times2)$ & conv1\_2-32$(5\times 5)$ + LRN\\ 
     LRN & {\color{red}dropin (1\_1 + 1\_2)} \\ \hline
     conv2-32$(5\times 5)$ & conv2\_1-32$(5\times 5)$ + LRN\\ 
     maxpool$(2\times2)$ & conv2\_2-32$(5\times 5)$ + LRN\\ 
     LRN    &					{\color{red}dropin (2\_1 + 2\_2)} \\ \hline
     & conv3\_1-32$(5\times 5)$ + LRN\\ 
     & conv3\_2-32$(5\times 5)$ + LRN\\ 
     &					{\color{red}dropin (3\_1 + 3\_2)} \\ \hline
     & conv4\_1-32$(5\times 5)$ + LRN\\ 
     & conv4\_2-32$(5\times 5)$ + LRN\\ 
     &					{\color{red}dropin (4\_1 + 4\_2)} \\ \hline
     & conv5\_1-32$(5\times 5)$ + LRN\\ 
     & conv5\_2-32$(5\times 5)$ + LRN\\ 
     &					{\color{red}dropin (5\_1 + 5\_2)} \\ \hline
     conv3-64$(3\times 3)$ & conv6\_1-64$(3\times 3)$ \\  \hline
\multicolumn{2}{|c|}{maxpool$(2\times2)$} \\ \hline
\multicolumn{2}{|c|}{fc-10} \\ \hline
\multicolumn{2}{|c|}{soft-max} \\ \hline
\multicolumn{2}{|c|}{accuracy} \\ \hline
  \end{tabular}
  \caption{CIFAR-10 11-layer architecture, including \dropin units.}
  \label{tab:cifarArch}
\end{center}
  \vspace{-25pt}
\end{table}

\begin{figure}[tb]
\begin{center}
   \includegraphics[width=1.0\linewidth]{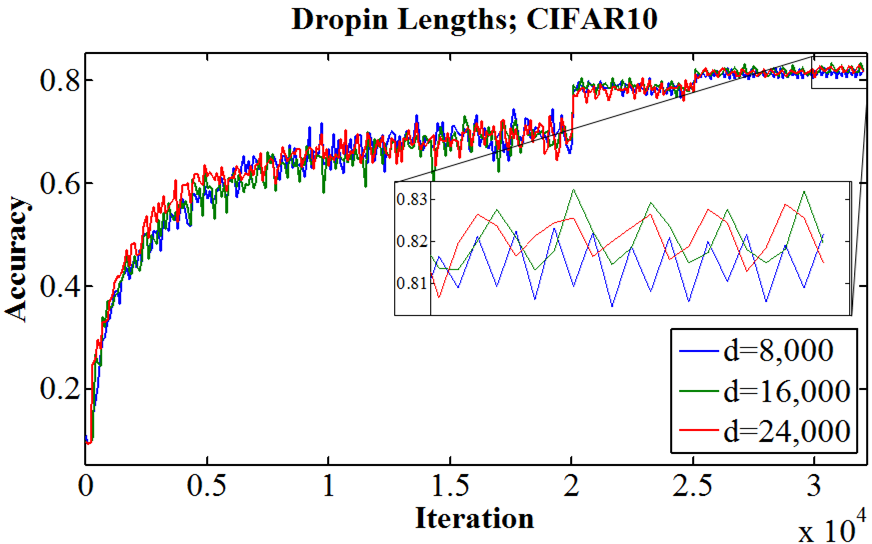}
\end{center}
  \vspace{-10pt}
   \caption{Test data classification accuracy while training the 11-layer CIFAR-10 architecture with \dropinNS. The curves show classification accuracies for different dropin\_lengths, $d$. (Best viewed in color)  
}
\label{fig:cifarDropinLen}
  \vspace{-5pt}
\end{figure}


\begin{table}[tb]
\begin{center}
  \begin{tabular}{| c | c | c |}
    \hline
    Architecture & dropin\_length & Accuracy (\%) \\ \hline \hline
   3-layer net &     & 81.4 \\ \hline
   11-layer net & 8,000  & 81.7 \\ \hline
   11-layer net & 16,000  & \textbf{82.3} \\ \hline
   11-layer net & 24,000  & \textbf{82.3} \\ \hline
  \end{tabular}
  \caption{Final accuracy (average of last three values) results for the CIFAR-10 dataset on test data at the end of the training.  Comparison of \dropin and dropin\_lengths. }
  \label{tab:cifarLengths}
\end{center}
  \vspace{-20pt}
\end{table}


\subsection{CIFAR-10} \label{sec:cifar}

This dataset consists of 60,000 color images with a resolution of 32x32.
Of these, 50,000 are for training and 10,000 are for testing.
There are ten classes with 6,000 images per class. 

The Caffe \cite{Caffe14} website provides the architecture and hyper-parameter settings as part of the CIFAR-10 tutorial\footnote{\url{http://caffe.berkeleyvision.org/gathered/examples/cifar10.html}}.
The three convolutional layer architecture trains quickly and attains good accuracies.
The convolutional layers were replicated to obtain an 11-layer model, which corresponds to the depth of one of the CIFAR-10 models in the experiments for highway networks \cite{srivastava2015training}. 
The detailed architectures are compared in Table \ref{tab:cifarArch}.
As shown in the table, the sizes of each of the layers entering the \dropin layer were kept the same for simplicity.
For every convolutional layer, the weight initialization was Gaussian with standard deviation of 0.01 and the bias initialization was constant, set to 0.
Each convolutional layer was followed by a rectified linear unit and local normalization.
The length of the training, the learning rates, and schedule were modified to run over 32,000 iterations.
This modification trained satisfactorily and provided a reasonable comparison.

Numerous attempts at training this 11-layer network without the \dropin layers failed to converge.
Similar attempts to train this network with the \dropin layers did successfully converge, which is a primary result of this study.

Experiments were performed varying the \dropin length.
Figure \ref{fig:cifarDropinLen} shows the accuracy curves for $ dropin\_length = 8,000, 16,000, 24,000$,  and Table \ref{tab:cifarLengths} compares the final accuracies.
The final accuracies show a marginal improvement for longer versus shorter lengths but for CIFAR-10 the results are relatively independent of the length value.
Furthermore, the final accuracies from the 11-layer architecture are less than 1\% better than the original 3-layer architecture, which implies that for the CIFAR-10 dataset, the deeper networker  provides  only marginal improvement.


\begin{table}[htb]
\begin{center}
  \begin{tabular}{| c | c |}
    \hline
     AlexNet  & AlexNet (13 layers) {\color{red}+ \dropin} \\ \hline
    
     \multicolumn{2}{|c|}{data $(227\times227\times3)$} \\ \hline
     conv1\_1-96$(11\times 11)$ & conv1\_1-96$(11\times 11)$ \\ 
      & conv1\_2-96$(11\times 11)$ \\ 
     &					{\color{red}dropin (1\_1 + 1\_2)} \\ \hline
	\multicolumn{2}{|c|}{maxpool$(2\times2)$ + LocalNorm} \\ \hline
     conv2\_1-256$(5\times 5)$ & conv2\_1-256$(5\times 5)$ \\ 
      & conv2\_2-256$(5\times 5)$ \\ 
     &					{\color{red}dropin (2\_1 + 2\_2)} \\ \hline
\multicolumn{2}{|c|}{maxpool$(2\times2)$ + LocalNorm} \\ \hline
     conv3\_1-384$(3\times 3)$ & conv3\_1-384$(3\times 3)$ \\ 
      & conv3\_2-384$(3\times 3)$ \\ 
     &					{\color{red}dropin (3\_1 + 3\_2)} \\ \hline
     conv4\_1-384$(3\times 3)$ & conv4\_1-384$(3\times 3)$ \\ 
      & conv4\_2-384$(3\times 3)$ \\
     &					{\color{red}dropin (4\_1 + 4\_2)} \\ \hline
     conv5\_1-256$(3\times 3)$ & conv5\_1-256$(3\times 3)$ \\ 
      & conv5\_2-256$(3\times 3)$ \\
     &					{\color{red}dropin (5\_1 + 5\_2)} \\ \hline
\multicolumn{2}{|c|}{maxpool$(2\times2)$} \\ \hline
\multicolumn{2}{|c|}{fc6-4096} \\ \hline
\multicolumn{2}{|c|}{fc7-4096} \\ \hline
\multicolumn{2}{|c|}{fc8-1000} \\ \hline
\multicolumn{2}{|c|}{soft-max} \\ \hline
  \end{tabular}
  \caption{Network architecture for AlexNet and modified version of AlexNet, AlexNet (13 layers) + \dropin.  }
  \label{tab:alexnet13}
\end{center}
  \vspace{-15pt}
\end{table}

\begin{figure}[htb]
\begin{center}
   \includegraphics[width=.901\linewidth]{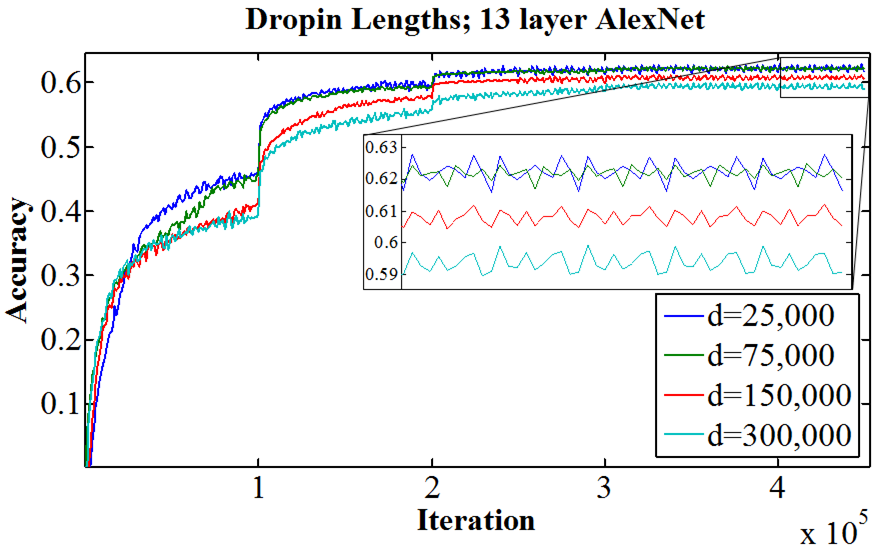}
\end{center}
  \vspace{-15pt}
   \caption{Comparison of various \dropin lengths, $d$.  Validation data classification accuracy while training the AlexNet (13 layers) + \dropin architecture with ImageNet data. (Best viewed in color) 
}
\label{fig:AlexLengths}
  \vspace{-5pt}
\end{figure}

\begin{table}[tb]
\begin{center}
  \begin{tabular}{| c | c | c |}
    \hline
    Architecture & dropin\_length & Accuracy (\%) \\ \hline \hline
   AlexNet &     & 58.0 \\ \hline
   13 layers + \dropin & 25,000  & \textbf{62.2} \\ \hline
   13 layers + \dropin & 75,000  & 62.1 \\ \hline
   13 layers + \dropin & 150,000  & 60.8 \\ \hline
   13 layers + \dropin & 300,000  & 59.3 \\ \hline
  \end{tabular}
  \caption{Comparison of \dropin and dropin\_lengths, $d$. The table shows final accuracy (average of last three values) results for the ImageNet dataset on validation data at the end of the training.}
  \label{tab:dropinLengths}
\end{center}
  \vspace{-20pt}
\end{table}


\subsection{ImageNet / AlexNet}
\label{sec:alexnet}

ImageNet\footnote{\url{www.image-net.org/}} \cite{Imagenet15} is a large image database based on the nouns in  WordNet.
This image database, used for the ImageNet Large Scale Visual Recognition Challenge, is commonly used as a basis of comparison in the deep learning literature.
The database contains 1.2 million training and 50,000 testing images covering 1,000 categories.

The Caffe website provides the architecture and hyper-parameter files for a slightly modified AlexNet. 
We downloaded the architecture and hyper-parameter files from the website and we expanded the architecture from 8 layers to 13 layers by duplicating each of the convolutional layers, which is shown (minus the \dropin layers shown in red) in columns 1 and 2, respectively, of Table \ref{tab:alexnet13}.
The AlexNet (13 layers) + \dropin includes a \dropin layer between every duplicated layer used to create AlexNet (13 layers).
Multiple attempts at training the AlexNet (13 layers) architecture in the conventional manner did not converge.
In the tests with the expanded architecture, the hyper-parameters were kept the same as provided by the Caffe website (even though our experiments with \dropin indicate that tuning them could improve the results).

Experiments were run varying the \dropin hyper-parameter \textit{dropin\_length}.
Table \ref{tab:dropinLengths} shows final accuracy results after training for 450,000 iterations with a range of lengths. 
Figure \ref{fig:AlexLengths} compares the accuracy during training of these experiments.
In contrast to the results with CIFAR-10, the \dropin length makes a difference with ImageNet.
We believe that this is because the deeper architecture increases the classification accuracy for larger datasets, hence  the improvement with smaller \dropin lengths is more prominent.

From Figure \ref{fig:AlexLengths} and Table \ref{tab:dropinLengths},  we can conclude that shorter lengths are better than the longer ones.
If the length is less than the first scheduled drop in the learning rate at iteration 100,000, then the network is better trained.
However, the difference between $dropin\_length = 75,000$ and 25,000 is negligible implying that lengths less than the first scheduled learning rate drop are equivalent.
\begin{table}[htb]
  \vspace{-5pt}
\begin{center}
  \begin{tabular}{| c | c |}
    \hline
     VGG8 & VGG16 {\color{red}+ DropIn} \\ \hline
    
     \multicolumn{2}{|c|}{data $(224\times224\times3)$} \\ \hline
      conv1\_1-64$(3\times 3)$ & conv1\_1-64$(3\times 3)$ \\ 
      								& conv1\_2-64$(3\times 3)$ \\ 
      								& {\color{red}dropin (1\_1 + 1\_2)}\\ \hline
	\multicolumn{2}{|c|}{maxpool$(2\times2)$} \\ \hline
      conv2\_1-128$(3\times 3)$ & conv2\_1-128$(3\times 3)$ \\ 
      & conv2\_2-128$(3\times 3)$ \\  
      & {\color{red}dropin (2\_1 + 2\_2)}\\ \hline
\multicolumn{2}{|c|}{maxpool$(2\times2)$} \\ \hline
     conv3\_1-256$(3\times 3)$ & conv3\_1-256$(3\times 3)$ \\ 
     &conv3\_2-256$(3\times 3)$ \\
     &  {\color{red}dropin (3\_1 + 3\_2)}\\ 
     &			 conv3\_3-256$(3\times 3)$ \\
     &		  {\color{red}dropin (3\_2 + 3\_3)}\\ \hline 
\multicolumn{2}{|c|}{maxpool$(2\times2)$} \\ \hline
      conv4\_1-512$(3\times 3)$& conv4\_1-512$(3\times 3)$ \\ 
      &conv4\_2-512$(3\times 3)$ \\
      & {\color{red}dropin (4\_1 + 4\_2)}\\ 
     &			 conv4\_3-512$(3\times 3)$ \\ 
     &  {\color{red}dropin (4\_2 + 4\_3)}\\ \hline
\multicolumn{2}{|c|}{maxpool$(2\times2)$} \\ \hline
     conv5\_1-512$(3\times 3)$ & conv5\_1-512$(3\times 3)$ \\ 
     &conv5\_2-512$(3\times 3)$ \\
 &{\color{red}dropin (5\_1 + 5\_2)}\\ 
  &   			 conv5\_3-512$(3\times 3)$ \\ 
   &  			      {\color{red} dropin (5\_2 + 5\_3)}\\ \hline
\multicolumn{2}{|c|}{maxpool$(2\times2)$} \\ \hline
\multicolumn{2}{|c|}{fc6-4096} \\ \hline
\multicolumn{2}{|c|}{fc7-4096} \\ \hline
\multicolumn{2}{|c|}{fc8-1000} \\ \hline
\multicolumn{2}{|c|}{soft-max} \\ \hline
    
  \end{tabular}
  \caption{Network architectures for VGG8 and VGG16 + \dropinNS.  See the text for additional settings.}
  \label{tab:vgg_dropin}
\end{center}
  \vspace{-15pt}
\end{table}

\begin{figure}[]
\begin{center}
   \includegraphics[width=1.0\linewidth]{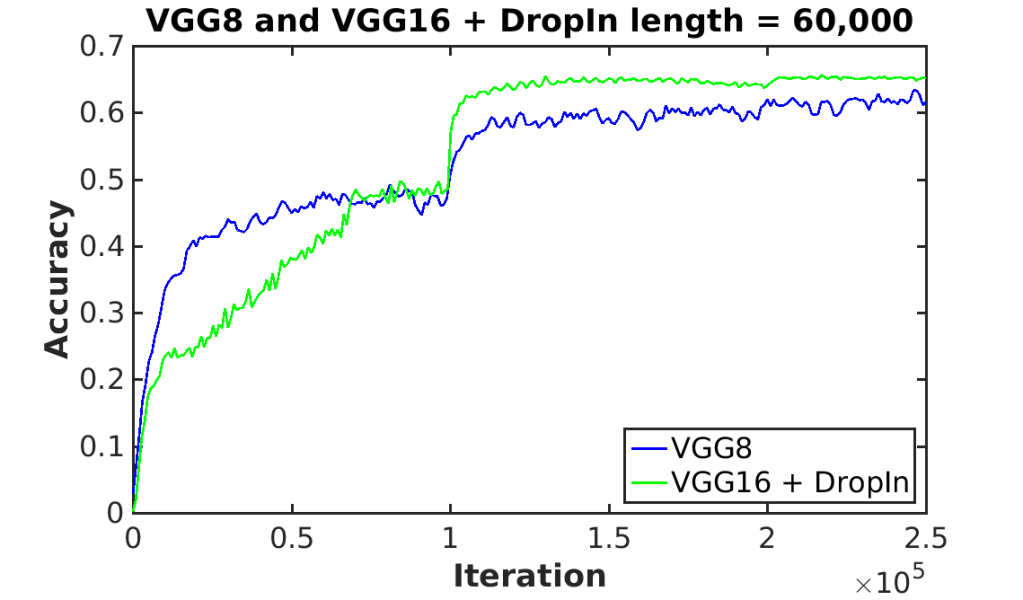}
\end{center}
  \vspace{-15pt}
   \caption{Validation data classification accuracy while training the VGG16 + \dropin architecture with ImageNet data. (Best viewed in color)
}
\label{fig:vgg_plot}
  \vspace{-10pt}
\end{figure}


\subsection{ImageNet / VGG}
\label{sec:vgg}
VGG$n$, a set of networks created by the Visual Geometry Group \cite{simonyan2014very}, won second place in the image classification category of the 2014 ImageNet contest.  These networks, trained on the same database as the Alexnet architecture discussed in Section \ref{sec:alexnet}, contained $ n=11,13,16,\mbox{ or }19$ layers.  In Table \ref{tab:vgg_dropin} we see the VGG16 (minus the \dropin layers shown in red) architecture alongside what we will refer to as VGG8 (not contained in the original paper).
All convolutional layers have a stride and padding of 1 and maxpooling layers have a stride of 2. 
In their paper, the authors describe the difficulty of training these deep networks and utilized a weight transfer method to enable the network to converge during training \cite{simonyan2014very}.


While it is possible to train a deep neural network by first training a shallow network and using those weights to initialize the deeper network, we believe that in addition to being easier, training the full network with all the layers in place leads to a better trained network.  This is supported by research on feature visualization, such as in Zeiler and Fergus \cite{zeiler2014visualizing}, where they demonstrate that higher layers have more abstract representations.
Training in place means that the learned representations will conform well to the representation at a given layer, while training a shallow network and initializing the weights of a deeper network might not.

Instead of training smaller networks, we propose to use our gradual DropIn method.  
For our studies, we utilized the VGG16 prototxt file referenced on the Caffe website\footnote{\url{https://gist.github.com/ksimonyan/211839e770f7b538e2d8\#file-vgg_ilsvrc_16_layers_deploy-prototxt}} 
and set up the solver file with the appropriate parameters from the authors' paper.  Using traditional training methods, we were only able to train the VGG8 architecture;  the VGG16 failed to begin converging for multiple realizations.  Using VGG8 as a template,  we augment VGG16 with \dropin layers to create VGG16 + \dropin (see Table \ref{tab:vgg_dropin}).  

Based on the evidence presented in Section \ref{sec:alexnet}, we choose to test VGG16 with a \dropin length of 60,000. We found that other lengths (100,000, 150,000, and 200,000) began to converge as well but with limited time and resources, we chose to report only this length for this paper.
The results of training VGG16 + \dropin are shown in Figure \ref{fig:vgg_plot} alongside VGG8.  We see that with gradual \dropin the difficult to train VGG16 network does converge.  Here we see the real power of the gradual \dropin method; without training an additional shallower network we are able to directly train VGG16,  thus saving effort for the practitioner. 


\begin{table}[h]
  \vspace{-5pt}
\begin{center}
  \begin{tabular}{| c | c | c |}
    \hline
    Case & fc6 & fc7 \\ \hline \hline
   1 & dropout  & dropout \\ \hline
   2 & dropout  &  \\ \hline
   3 & dropout  & \dropin \\ \hline
  \end{tabular}
  \caption{The three regularization experiments shows layers with dropout or \dropin. The fully connected layers 6 and 7, are called \textit{fc6} and \textit{fc7}, respectively. }
  \label{tab:reg}
\end{center}
  \vspace{-15pt}
\end{table}

\begin{figure}[h]
\begin{center}
   \includegraphics[width=.95\linewidth]{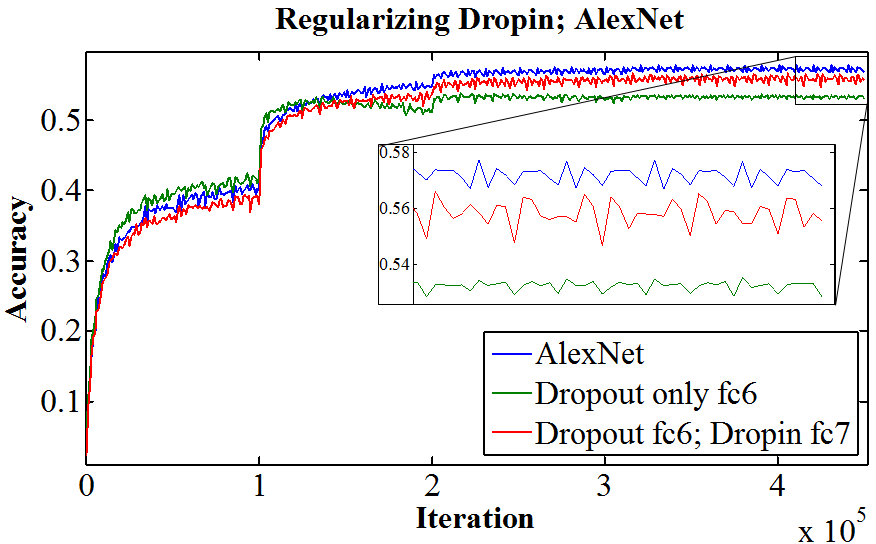}
\end{center}
  \vspace{-10pt}
   \caption{Test of \dropin regularization with AlexNet. Validation data classification accuracy while training AlexNet with ImageNet data.  
(Best viewed in color) 
}
\label{fig:regularization}
  \vspace{-15pt}
\end{figure}


\subsection{Using \dropin for regularization}
\label{sec:regularization}
The original AlexNet architecture uses dropout for regularization during training in both fully connected layers and it provides a substantial increase in the network's accuracy.
AlexNet (with 8 layers) provides a means to test \dropin regularization.
For this experiment, three cases were run as shown in  Table \ref{tab:reg}.
Case 1 is the original AlexNet.


The results from this experiment are shown in Figure \ref{fig:regularization},
where both \dropin and dropout probability ratios were $ 0.5 $ for all of these tests and all the other hyper-parameters were the same.
This figure shows that removing dropout from fc7 causes visible degrading of the accuracy between iterations 150,000 and 200,000 (green curve).
This kind of degradation does not happen with \dropinNS.
Instead, the accuracy curve is similar to the curve with dropout (red versus blue curve) but with a small degradation in overall performance.
We believe this degradation is because a \dropin network is more difficult to train than a dropout network.
However, the final accuracy for the network with \dropin in fc7 is higher than from an architecture without dropout (red versus green curve).
This experiment  demonstrates that \dropin provides some regularization since the degradation found in the case without dropout is absent.



\section{How to determine a good architecture}

One of the challenges for deep learning practitioners is to determine good choices for the hyper-parameter values and the architecture for a given application and dataset.
\dropin and dropout provide an easier way to test choices for the architecture than running a set of experiments with many different architectures.

\dropin and dropout can allow one to test a range of architecture depths and widths, respectively.
Since adding layers does not necessarily increase accuracy, one can run with the gradual \dropin mode to see if there is little effect, such as in Figures \ref{fig:lenet_10} and  \ref{fig:cifarDropinLen}, or visible effect, such as in Figure \ref{fig:AlexLengths}.
Substantial improvement implies that there will be benefit from the additional depth.

Similarly, making a run where the dropout ratio varies from perhaps 0.9 to 0.1 (using a slightly modified dropout) provides guidance on the minimum number of neurons per layer.
When decreasing the probability that neurons are retained (as shown in Figure 9 of Srivastava \etal \cite{srivastava2014dropout}), the error typically has a range of the probability ratios where the error plateaus but at some threshold probability the error increases.
By multiplying the number of neurons in a layer by this threshold probability, one can approximately determine the minimum number of neurons one must retain where there is marginal harm to the accuracy.




\section{Conclusion}

The major result of this paper is that deeper architectures that cannot converge using standard training methods, become trainable by slowly adding in the new layers during the training.  
In addition, there are indications that  \dropin layers help regularize the training of a network. 
We found in general that if the shallow network is trainable, then the deeper network, where  additional layers are added by a \dropin layer, is also trainable.  
With a large dataset like ImageNet, adding additional layers increases accuracy.

We have not yet explored training with tailored \dropin lengths for different \dropin layers in a network.
In addition, comparing \dropin to initializing the weights from training a separate shallow network has not yet been tested; these are planned for future work and will be reported elsewhere.  
Also we plan to test \dropin within other architectures such as recurrent neural networks.
Future work also includes training networks with hundreds of layers using asynchronous \dropinNS, where layers are added starting at different iterations.
In addition, we wish to test training where the entire very deep network is initially very thin (few parameters to train) and units are added to all the layers during the training.
Furthermore, we plan to study if a methodology can be developed to learn from the data how to automatically optimize the architecture during training and thus learn to adapt to an application based on its data.


{\small
\bibliographystyle{ieee}
\bibliography{dropin.bib}

\begin{thebibliography}{10}\itemsep=-1pt

\bibitem{baldi2013understanding}
P.~Baldi and P.~J. Sadowski.
\newblock Understanding dropout.
\newblock In {\em Advances in Neural Information Processing Systems}, pages
  2814--2822, 2013.

\bibitem{bengio2009curriculum}
Y.~Bengio, J.~Louradour, R.~Collobert, and J.~Weston.
\newblock Curriculum learning.
\newblock In {\em Proceedings of the 26th Annual International Conference on
  Machine Learning}, pages 41--48. ACM, 2009.

\bibitem{breuel2015possible}
T.~M. Breuel.
\newblock Possible mechanisms for neural reconfigurability and their
  implications.
\newblock {\em arXiv preprint arXiv:1508.02792}, 2015.

\bibitem{erhan2009difficulty}
D.~Erhan, P.-A. Manzagol, Y.~Bengio, S.~Bengio, and P.~Vincent.
\newblock The difficulty of training deep architectures and the effect of
  unsupervised pre-training.
\newblock In {\em International Conference on Artificial Intelligence and
  Statistics}, pages 153--160, 2009.

\bibitem{girshick2014rich}
R.~Girshick, J.~Donahue, T.~Darrell, and J.~Malik.
\newblock Rich feature hierarchies for accurate object detection and semantic
  segmentation.
\newblock In {\em Computer Vision and Pattern Recognition (CVPR), 2014 IEEE
  Conference on}, pages 580--587. IEEE, 2014.

\bibitem{glorot2010understanding}
X.~Glorot and Y.~Bengio.
\newblock Understanding the difficulty of training deep feedforward neural
  networks.
\newblock In {\em International conference on artificial intelligence and
  statistics}, pages 249--256, 2010.

\bibitem{graves2014towards}
A.~Graves and N.~Jaitly.
\newblock Towards end-to-end speech recognition with recurrent neural networks.
\newblock In {\em Proceedings of the 31st International Conference on Machine
  Learning (ICML-14)}, pages 1764--1772, 2014.

\bibitem{he2015delving}
K.~He, X.~Zhang, S.~Ren, and J.~Sun.
\newblock Delving deep into rectifiers: Surpassing human-level performance on
  imagenet classification.
\newblock {\em arXiv preprint arXiv:1502.01852}, 2015.

\bibitem{hinton2006fast}
G.~E. Hinton, S.~Osindero, and Y.-W. Teh.
\newblock A fast learning algorithm for deep belief nets.
\newblock {\em Neural Computation}, 18(7):1527--1554, 2006.

\bibitem{hinton2012improving}
G.~E. Hinton, N.~Srivastava, A.~Krizhevsky, I.~Sutskever, and R.~R.
  Salakhutdinov.
\newblock Improving neural networks by preventing co-adaptation of feature
  detectors.
\newblock {\em arXiv preprint arXiv:1207.0580}, 2012.

\bibitem{huval2015empirical}
B.~Huval, T.~Wang, S.~Tandon, J.~Kiske, W.~Song, J.~Pazhayampallil,
  M.~Andriluka, R.~Cheng-Yue, F.~Mujica, A.~Coates, et~al.
\newblock An empirical evaluation of deep learning on highway driving.
\newblock {\em arXiv preprint arXiv:1504.01716}, 2015.

\bibitem{Caffe14}
Y.~Jia, E.~Shelhamer, J.~Donahue, S.~Karayev, J.~Long, R.~Girshick,
  S.~Guadarrama, and T.~Darrell.
\newblock Caffe: Convolutional architecture for fast feature embedding.
\newblock {\em In Proceedings of the ACM International Conference on
  Multimedia}, pages 675--678, 2014.

\bibitem{Alexnet12}
A.~Krizhevsky, I.~Sutskever, and G.~E. Hinton.
\newblock {Imagenet classification with deep convolutional neural networks}.
\newblock {\em Advances in Neural Information Processing Systems}, 2012.

\bibitem{lecun2015deep}
Y.~LeCun, Y.~Bengio, and G.~Hinton.
\newblock Deep learning.
\newblock {\em Nature}, 521(7553):436--444, 2015.

\bibitem{montufar2014number}
G.~F. Montufar, R.~Pascanu, K.~Cho, and Y.~Bengio.
\newblock On the number of linear regions of deep neural networks.
\newblock In {\em Advances in Neural Information Processing Systems}, pages
  2924--2932, 2014.

\bibitem{raiko2012deep}
T.~Raiko, H.~Valpola, and Y.~LeCun.
\newblock Deep learning made easier by linear transformations in perceptrons.
\newblock In {\em International Conference on Artificial Intelligence and
  Statistics}, pages 924--932, 2012.

\bibitem{romero2014fitnets}
A.~Romero, N.~Ballas, S.~E. Kahou, A.~Chassang, C.~Gatta, and Y.~Bengio.
\newblock Fitnets: Hints for thin deep nets.
\newblock {\em arXiv preprint arXiv:1412.6550}, 2014.

\bibitem{Imagenet15}
O.~Russakovsky, J.~Deng, H.~Su, J.~Krause, S.~Satheesh, S.~Ma, Z.~Huang,
  A.~Karpathy, A.~Khosla, M.~Bernstein, A.~C. Berg, and L.~Fei-Fei.
\newblock {ImageNet Large Scale Visual Recognition Challenge}.
\newblock {\em International Journal of Computer Vision (IJCV)}, 2015.

\bibitem{simonyan2014very}
K.~Simonyan and A.~Zisserman.
\newblock Very deep convolutional networks for large-scale image recognition.
\newblock {\em arXiv preprint arXiv:1409.1556}, 2014.

\bibitem{srivastava2014dropout}
N.~Srivastava, G.~Hinton, A.~Krizhevsky, I.~Sutskever, and R.~Salakhutdinov.
\newblock Dropout: A simple way to prevent neural networks from overfitting.
\newblock {\em The Journal of Machine Learning Research}, 15(1):1929--1958,
  2014.

\bibitem{srivastava2015highway}
R.~K. Srivastava, K.~Greff, and J.~Schmidhuber.
\newblock Highway networks.
\newblock {\em arXiv preprint arXiv:1505.00387}, 2015.

\bibitem{srivastava2015training}
R.~K. Srivastava, K.~Greff, and J.~Schmidhuber.
\newblock Training very deep networks.
\newblock {\em arXiv preprint arXiv:1507.06228}, 2015.

\bibitem{sussillo2015random}
D.~Sussillo and L.~Abbott.
\newblock Random walk initialization for training very deep feedforward
  networks.
\newblock {\em arXiv preprint arXiv:1412.6558}, 2015.

\bibitem{sutskever2013importance}
I.~Sutskever, J.~Martens, G.~Dahl, and G.~Hinton.
\newblock On the importance of initialization and momentum in deep learning.
\newblock In {\em Proceedings of the 30th International Conference on Machine
  Learning (ICML-13)}, pages 1139--1147, 2013.

\bibitem{sutskever2014sequence}
I.~Sutskever, O.~Vinyals, and Q.~V. Le.
\newblock Sequence to sequence learning with neural networks.
\newblock In {\em Advances in Neural Information Processing Systems}, pages
  3104--3112, 2014.

\bibitem{szegedy2014going}
C.~Szegedy, W.~Liu, Y.~Jia, P.~Sermanet, S.~Reed, D.~Anguelov, D.~Erhan,
  V.~Vanhoucke, and A.~Rabinovich.
\newblock Going deeper with convolutions.
\newblock {\em arXiv preprint arXiv:1409.4842}, 2014.

\bibitem{taigman2014deepface}
Y.~Taigman, M.~Yang, M.~Ranzato, and L.~Wolf.
\newblock Deepface: Closing the gap to human-level performance in face
  verification.
\newblock In {\em Computer Vision and Pattern Recognition (CVPR), 2014 IEEE
  Conference on}, pages 1701--1708. IEEE, 2014.

\bibitem{vinyals2014show}
O.~Vinyals, A.~Toshev, S.~Bengio, and D.~Erhan.
\newblock Show and tell: A neural image caption generator.
\newblock {\em arXiv preprint arXiv:1411.4555}, 2014.

\bibitem{zeiler2014visualizing}
M.~D. Zeiler and R.~Fergus.
\newblock Visualizing and understanding convolutional networks.
\newblock In {\em Computer Vision--ECCV 2014}, pages 818--833. Springer, 2014.

\bibitem{Zhang2015highway}
Y.~Zhang, G.~Chen, D.~Yu, K.~Yao, S.~Khudanpur, and J.~Glass.
\newblock Highway long short-term memory rnns for distant speech recognition.
\newblock {\em arXiv preprint arXiv:1510.08983}, 2015.

\end{thebibliography}
}

\end{document}